\documentclass[runningheads]{llncs}
\usepackage{graphicx}
\usepackage{amsmath}
\usepackage[ruled,vlined]{algorithm2e}
\usepackage[dvipsnames]{xcolor}

\definecolor{mypink1}{rgb}{0.858, 0.188, 0.478}
\begin{document}
\title{Coarse-to-Fine Object Tracking Using Deep Features and Correlation filters}
%
%

\author{Ahmed Zgaren\inst{1} \and Wassim Bouachir\inst{1} \and Riadh Ksantini\inst{2}}
 \authorrunning{A. Zgaren et al.}
%
 \institute{\textsuperscript{1}Department of Science and Technology, TELUQ University\\
 \email{\{ahmed.zgaren, wassim.bouachir\}@teluq.ca}\\
 \textsuperscript{2}Department of Computer Science, University of Bahrain\\
 \email{rksantini@uob.edu.bh}}

\maketitle              
\begin{abstract}
During the last years, deep learning trackers achieved stimulating results while bringing interesting ideas to solve the tracking problem. This progress is mainly due to the use of learned deep features obtained by training deep convolutional neural networks (CNNs) on large image databases. But since CNNs were originally developed for image classification, appearance modeling provided by their deep layers might be not enough discriminative for the tracking task. In fact, such features represent high-level information, that is more related to object category than to a specific instance of the object. Motivated by this observation, and by the fact that discriminative correlation filters (DCFs) may provide a complimentary low-level information, we present a novel tracking algorithm taking advantage of both approaches. We formulate the tracking task as a two-stage procedure. First, we exploit the generalization ability of deep features to coarsely estimate target translation, while ensuring invariance to appearance change. Then, we capitalize on the discriminative power of correlation filters to precisely localize the tracked object. Furthermore, we designed an update control mechanism to learn appearance change while avoiding model drift. We evaluated the proposed tracker on object tracking benchmarks. Experimental results show the robustness of our algorithm, which performs favorably against CNN and DCF-based trackers. Code is available at:  \texttt{\color{mypink1} \textit{https://github.com/AhmedZgaren/Coarse-to-fine-Tracker}}

\end{abstract}
\section{Introduction}
 
The success of CNNs for visual object tracking (VOT) is mainly attributed to their rich feature hierarchy, and to the ability of convolutional layers to provide invariant feature representation against target appearance change. However, a major limitation relies in the fact that CNNs were developed following the principals of other visual classification tasks, where the goal is to predict a class label. They were thus designed without considering fundamental differences with the tracking task, that aims to locate a target object on an image sequence. Adopting such models for tracking may result in using general and redundant target representations, that are not enough discriminative for the tracking problem. For example, many trackers use feature representations from deep layers \cite{Wang2020TrackingBI,siamrpn,Danelljan_2019_CVPR,GCT,Wang_2019_CVPR,Chi2017DualDN}. These layers are naturally related to object category semantics and do not consider object specificities and intra-category variation. In the tracking task, object specific characteristics are important in order to determine its precise location on the image, and especially for distinguishing between the target and distractors (i.e. objects belonging to the same class and thus having similar appearance).

 Some authors attempted to address this issue by  combining feature representations provided by different CNN layers \cite{Ma2015HierarchicalCF,Ma2016WhenCF,Qi2016HedgedDT,ccot,touil2019hierarchical}. Such algorithms exploit multiple layers for feature extraction, based on the idea that different layers in a CNN model provide several detail levels in characterizing an object. Moreover, most of these methods incorporate adaptive correlation filters learned on multiple CNN layers, which is shown to improve tracking precision \cite{Ma2015HierarchicalCF}.
Our work proceeds along the same direction, in that it aims to investigate the optimal exploitation of deep features and correlation filters in order to improve tracking robustness. However, our approach differs from previous works by a new formulation of the VOT task, proposing a different way for combining CNNs and DCFs within the tracking framework. We formulate target search on a video frame as a two-stage procedure using deep features and DCFs sequentially. Since the last convolutional layers provide features that are related to object category semantics, we firstly exploit such representations to perform a coarse estimation of the target translation, while ensuring robustness against appearance change. Then, a precise localization is performed by applying a learned DCF filter and selecting the maximum correlation response within the coarse region. Unlike feature maps from deep layers, DCF filters are able to capture spatial details  related to object specific characteristics, and thus provide low-level features that are important to determine a precise location.

\begin{figure}
    \centering
    \includegraphics[width=\textwidth]{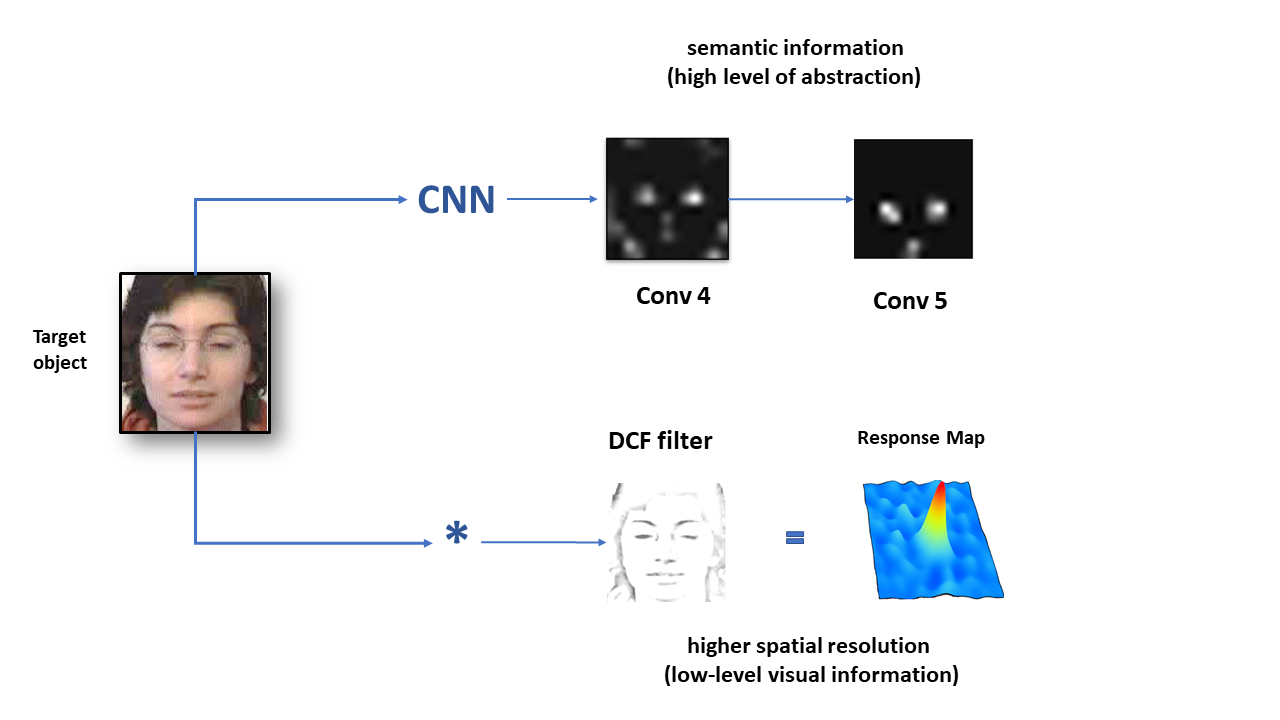}
    \caption{illustration of our two feature extraction levels. Top: high-level semantic information can be extracted from deep convolutional layers (e.g. conv4 and conv5 from AlexNet). Bottom: DCF filters produce response maps corresponding to low-level spatial information.}
    \label{fig:1}
\end{figure}

In figure \ref{fig:1}, we consider an example of a target face to emphasize the two feature levels that motivated our method. We can observe that CNN feature maps corresponding to deep convolutional layers encode the face appearance in a summarized fashion. This illustrates the generalization ability of such representations and their robustness to appearance change. But since these deep layers are naturally related to object category semantics, they do not guarantee precise localisation, nor robustness against distractors. On the contrary, the second row of figure \ref{fig:1} shows that DCF filters provide low-level visual information. This information is more related to object specific characteristics, which makes DCFs more appropriate for precise localisation of the target, and for handling tracking difficulties, such as the presence of distractors (e.g. tracking a face or a pedestrian in a crowded scene).     

Once the new target position is predicted, we update the appearance model to learn appearance variations of the object. Handling appearance change is among the main challenges in VOT. Most DCF-based trackers (e.g. \cite{srdcf,ccot}) are continuously updated on-the-fly, at a risk of contaminating the model by inappropriate updates. On the contrary, certain deep trackers use target features extracted only from the first frame or the previous frame \cite{siamfc,he2018twofold}, which  is known to speed-up tracking while reducing accuracy. We believe that a dynamic model that evolves during the tracking is important to handle appearance variations. However, we argue that excessive updates may contaminate the model due to an over-fitting to sudden variations. We therefore propose an update control mechanism to determine if the tracking status is appropriate for learning new feature representations. Such a mechanism makes it possible to optimize online appearance learning with respect to perturbation factors that may contaminate the model (e.g. occlusion).

To sum up, the main contributions of our work are as follows. First, we propose an effective coarse-to-fine approach to exploit deep features and correlation filters for object tracking. Second, to handle appearance change, we propose an update control mechanism, which allows learning new features during tracking, while avoiding model contamination. Third, we present extensive experiments on standard tracking benchmarks. The experimental work includes an ablation analysis to evaluate different design choices, as well as a comparative evaluation showing an improved performance compared to state-of-the-art trackers.  

\section{Related work}
\subsection{CNN for visual tracking}

Deep learning trackers typically use neural networks according to two main approaches:
\begin{enumerate}
	\item As a feature extractor \cite{eco,cui2016}, by extracting the features produced by a single layer or multiple layers for appearance modeling. Target search is then performed using traditional methods.
	\item For both feature extraction and target search, which is referred as end-to-end tracking \cite{Li2018MultiBranchSN,Wang2020TrackingBI}. In this case, target states are evaluated using the network output, that may have various forms such as probability maps and object classification scores.
\end{enumerate}

An important limitation of many deep trackers is related to the fact that target localization mainly depends on features from the last convolutional layers. Since CNN models were initially developed for image classification, deep feature maps provide high-level information that is more related to an object class, than to a specific instance of an object. In our work, we argue that deep features are naturally more appropriate to coarsely estimate target translation during a preliminary prediction step of the tracking process.

\subsection{Tracking with discriminant correlation filters}
The main idea of DCF tracking is to initially learn a filter from the target image region on the first frame. Then for each subsequent frame, the filter is used as a linear classifier to compute correlation over a search window, and discriminate between the target and the background. The new target position is predicted as the maximum value in the correlation output. Since the pioneering work of MOSSE \cite{mosse}, DCF trackers have been extensively studied as an efficient solution for the tracking problem. significant improvements have been then released on the DCF framework to address inherent limitations. For example, Henriques et al. \cite{kcf,Henriques2012ExploitingTC} proposed to incorporate multiple channels and kernels to allow the non-linear classification of boundaries. Further, Daneljann et al. \cite{srdcf} proposed the Spatially Regularized Discriminative Correlation Filters (SRDCF). They introduced a spatial regularization component within the DCF tracking framework to handle the periodic assumption problems. The proposed regularization weights penalize the CF coefficients, to allow learning the filter  on larger image regions. In addition, the DCF framework has been enhanced by, including scale estimation \cite{Danelljan2014AccurateSE} and a long-term memory \cite{Ma2015LongtermCT}.

Despite major amelioration, DCF based trackers still suffer from high sensitivity to appearance change (e.g. object deformation, motion blur). This limitation is mainly due to their high spatial resolution, which limits their ability to generalize and learn semantic information on the target object. Another limitation relies to their continuous learning procedure, at the risk of contaminating the model by inappropriate update (e.g. occlusion). Our tracking framework takes advantage of the high spatial resolution of DCFs for precise localization of the target, while exploiting the high level of abstraction provided by deep CNN layers to ensure robustness to appearance change. Moreover, we handle the online learning problem by using a control mechanism to avoid inappropriate updates. 

\subsection{Combining CNNs and DCFs for object tracking}
Existing CNN-based DCFs mainly focus on integrating convolutional features learned on deep network layers \cite{Ma2015HierarchicalCF,deepsrdcf,ccot,eco,Qi2016HedgedDT}. Ma et al. \cite{Ma2015HierarchicalCF} propose a hierarchical ensemble method of independent DCF trackers to combine convolutional layers. Qi et al. \cite{Qi2016HedgedDT} learn a correlation filter for each feature map, and use a modified Hedge algorithm to combine predictions. DeepSRDCF \cite{deepsrdcf} investigates the use of features extracted from the first convolutional layer of a CNN to train a correlation filter for tracking. The C-COT \cite{ccot} framework learns a convolutional operator in the continuous spatial domain using a pre-trained CNN, while ECO \cite{eco} proposes a factorized formulation to avoid the over-fitting problem observed with C-COT \cite{ccot}, and to reduce the number of learned filters. 

In our work, we propose a different way to incorporate CNNs and DCFs within the tracking framework, in order to ensure the optimal exploitation of both models. We decompose target search into two stages to exploit the power of CNNs for high-level feature extraction, and the high accuracy of DCFs for precise target localization.

\section{Proposed method}

\begin{figure}
\centering
\includegraphics[width=\textwidth]{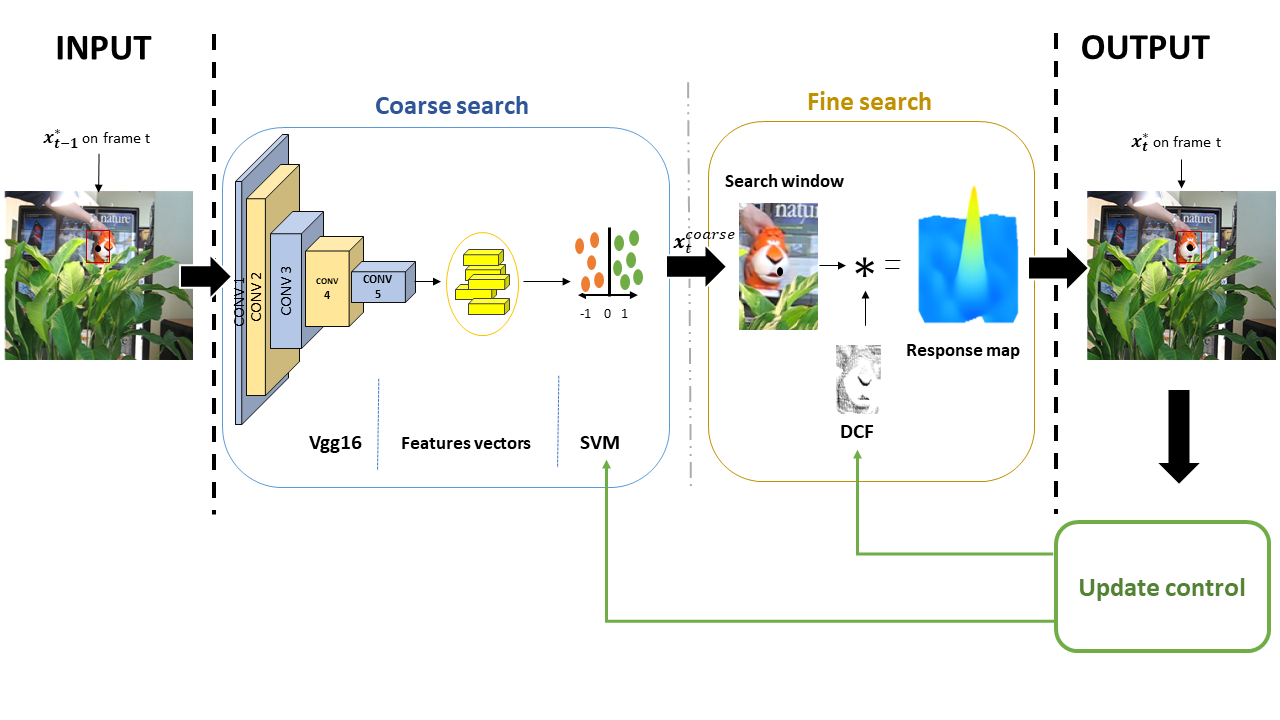}
\caption{The architecture of our coarse-to-fine tracker.}
\label{fig:2}
\end{figure}

We propose a coarse-to-fine tracking approach by combining CNNs with DCFs as two different, yet complementary, feature levels. In our framework, the correlation operation within the DCF acts similarly to a CNN convolutional layer. In fact, the learned correlation filter can be viewed as a final classification layer in the neural network. Thus, the  tracker takes advantage of both models, while overcoming their limitations mutually. In this section, we decompose the tracking procedure into three steps: (1) coarse target search by using an incremental SVM fed with deep features, (2) fine target prediction as the maximum correlation response of a learned DCF within the coarse region, and (3) adaptation to appearance change through an update control mechanism. Figure \ref{fig:2} illustrates the main steps of our algorithm.

\subsection{ Convolutional features for coarse search}

We exploit deep convolutional layers of CNN to ensure representations that are robust to appearance variation and deformation. Our conception suggests to encode object information using a feature vector $T$ that summarises the output of the $K$ activation maps forming a convolutional layer. For dimensionality reduction, each activation map $\mathcal{M}_h$,  $h \in \{1,2,..., K\}$  is up-sampled to produce a feature vector $T$ as:

\begin{equation}
    T(h)\:=\:\sum_{i=1}^{m}\:\sum_{j=1}^{n} \:\mathcal{M}_h(i,j) \lambda\:,
\label{eq:1}
\end{equation}
where $m \times n$ is the spatial resolution of the feature map, and $\lambda$ is a regularization parameter. 

At the first frame, the feature vector $T$ is extracted from target region using a pre-trained CNN. Since target information extracted from the first frame is initially insufficient for appearance modeling, we use data augmentation to generate additional learning examples through rotation, scale variation, and translation. The corresponding feature vectors are then used to feed a machine learning model used for identifying region proposals that are likely to be the target.
During each tracking iteration, we evaluate several candidate regions in a search window centered around the last target location $\textbf{x}^{*}_{t-1}$. Candidate regions are first generated by varying  polar coordinates, with respect to the last target location center. 
Given the feature vector $T_i$ of the $i$th candidate region, we compute the corresponding weight $P_i$ ($P_i \in [-1,1] $) using an incremental one-class SVM \cite{kefi2019novel} learned from previous frames. $P_i$  represents the likelihood of the candidate region of being the target, and is calculated using the SVM classification function:
\begin{equation}
P_i = \Delta_{W,b}(T_i).
\label{eq:2}
\end{equation}
Here, $W$ and $b$ denote the weight vector and the bias of the SVM, respectively.

The position  $\textbf{x}^{coarse}_t$ corresponding to the center of the coarse search region is finally estimated as the average weighted position over the best $q$ candidate regions (based on their likelihoods), as follows:
\begin{equation}
 \textbf{x}^{coarse}_t =\frac{\sum_{i=1}^{q}{P_i \:\textbf{x}_i}}{\sum_{i=1}^q{P_i}}, 
 \label{eq:3}
\end{equation}
where  $\textbf{x}_i$ is the center of the $i$th candidate region. The output of this step is considered as a coarse estimation of the target translation between two consecutive frames.

\subsection{Fine search}

The second stage of our tracking procedure is based on DCF to find the maximum correlation output in a search region centered around $\textbf{x}^{coarse}_t$. Since DCF trackers assume that the target undergoes small displacements between two consecutive frames, the search region at frame $t$ is classically defined as a sub-window centered around the last predicted position. Instead, we define the correlation search region as the sub-window $r_t$ of size $l\: \times \: z$, centered around $\textbf{x}^{coarse}_t$. In this manner, we relax the small displacement assumption and rely on the coarse estimation of the target translation to determine the target search area.    

The search sub-window $r_t$ is extracted and augmented periodically. The filter $f_{t-1}$ is then applied in a sliding window-like manner. The response map $Y_{f}(r_t)$ is constructed from the correlation scores at each position, as the inner product between the filter $f_{t-1}$ and the shifted sub-image at that position. The correlation is computed in the Fourier domain, to produce the correlation output map $Y_{f}(r_t)$ using the convolution property of the FFT, that is: 

\begin{equation}
Y_{f}{(r_t)}=F^{-1}\Big( \:F(f_{t-1})\: \odot F(r_t)\:\Big),
\label{eq:4}
\end{equation}
where $\odot$ denotes point-wise multiplication, and $F^{-1}$ the inverse FFT.

We note that the filter is applied at multiple resolutions to estimate target scale changes. Generally, the output response map approximately follows a Gaussian distribution, as CF-based trackers are trained with Gaussian shaped regression labels. The final position $\textbf{x}^*_t$ of the target on frame $t$ corresponds to the maximal correlation response calculated on all cyclic shifts of region $r_t$.

\subsection{Model Update}

During tracking, DCF-based trackers are typically updated on-the-fly at a risk of contaminating the model due to over-fitting to sudden changes or other inappropriate update situations such as occlusion. In order to optimize appearance modeling with respect to such perturbation factors, we use an update control mechanism to determine if the tracking status is appropriate for learning new feature representations.  In particular, the feature vector $T^*_t$ is extracted from the predicted target region (using Eq. \ref{eq:1}), and a quality indicator $I_t$ is calculated using the SVM score function:
\begin{equation}
    I_t = \Delta_{W,b}{(T^{*}_t)}.
    \label{eq:6}
\end{equation}
Selective update is performed for each appearance model separately, if $I_t$ exceeds the corresponding minimum quality threshold. That is to say, two minimum quality thresholds $\mu$ and $\gamma$ are considered for the SVM and the DCF, respectively.   

On the one hand, SVM adaptation is carried out by exploiting the incremental property of the model to incorporate the new observation, as stated in \cite{kefi2019novel}. On the other hand, we update the filter according to a learning rate $\beta$ as follows:   
\begin{equation}
   F( f_t ) = \beta\: F( f_t ) + (1 - \beta)\: F( f_{t-1}), 
   \label{eq:5}
\end{equation}
with $f_t$ and $f_{t-1}$ denoting the filter at frames $t$ and $t-1$ respectively. The entire tracking process in summarized in algorithm \ref{alg:1}.

\begin{algorithm}[H]
\SetAlgoLined
\KwResult{Current position of the target $\textbf{x}^*_t$}

    extract $T$ from the first frame using Eq (\ref{eq:1})\;
    learn classifier $\Delta_{W , b}$ \;
    learn filter $f_0$ from the first frame\;
    
\For{each subsequent frame $t$}{
  generate candidate regions around $\textbf{x}^*_{t-1}$\;
  \For{each \textbf{i}th candidate region}{
    extract $T_i$ using Eq (\ref{eq:1})\;
    calculate $P_i$ using Eq (\ref{eq:2})\;
  }
  Compute $\textbf{x}^{coarse}_{t}$ using Eq (\ref{eq:3}) \;
  Calculate correlation map using Eq (\ref{eq:4})\;
  Select $\textbf{x}^*_t$ as the maximum correlation response\;
  calculate $I_t$ using Eq (\ref{eq:6})\;
  \If{$I_t \geq \mu$}{
   update $\Delta_{W , b}$\;
   }
  \If{$I_t \geq \gamma$ }{
   update $f_t$ using Eq (\ref{eq:5})\;
  }
}
 \caption{Tracking}
\label{alg:1}
\end{algorithm}

\section{Experiments}
\subsection{Implementation details}
For extracting deep features, we adopted the VGG16 network \cite{vggnet} trained on ImageNet \cite{ImageNet}. More specifically, we used the output of the convolutional layer conv5-3, which produces $K = 512$ feature maps with a spatial size $m \times n$ equal to $14  \times 14$. The feature vector $T$ is constructed by setting the regularization parameter $\lambda$ to 0.1. For the iCOSVM \cite{kefi2019novel}, we set the internal training parameter $\nu$ to 0.1 and the update threshold $\mu$ to 0.4. Our implementation of the fine search step is based on SRDCF, where we use the same parameters in \cite{srdcf} for training the correlation filter. We set the search region $l \times z$ to four times the target size and the learning rate $\beta$ to 0.025. We also fixed the update parameter $\gamma$ to 0.0 (recall that $I_t \in [-1,1]$). The proposed Framework is implemented using Matlab on a PC with a Intel i7-8700 3.2 GHz CPU and a Nvidia GeForce GTX 1070 Ti GPU.

\subsection{Evaluation methodology}
We performed a comprehensive evaluation of the proposed method on the standard benchmarks OTB100 \cite{Wu2015ObjectTB} and OTB50 \cite{Wu2013OnlineOT}. First, we present an ablation study to evaluate the importance of certain tracking mechanisms individually. We then compare our tracker to several state-of-the-art algorithms, including DCF-based trackers (SRDCF \cite{srdcf}, DSST \cite{Danelljan2014AccurateSE}, KCF\cite{kcf},CSK \cite{Henriques2012ExploitingTC}, and SAMF\cite{samf}), and CNN-based trackers (CNN-SVM \cite{cnnsvm}, DLT \cite{wang2013}, and SiamFC \cite{siamfc}). We follow the evaluation protocol presented in the benchmarks \cite{Wu2015ObjectTB} and  \cite{Wu2013OnlineOT}. Tracking performance is evaluated based on (1) the bounding box overlap ratio, and (2) the center location error. The corresponding results are visualized by the success and precision plots, respectively. These plots are generated by calculating ratios of successful tracking iterations at several thresholds. The Area Under Curve (AUC) scores are used to rank the compared tracking methods in the success plot. In the precision plots, the final ranks depend on the tracking accuracy at a threshold of 20 pixels. Note that all the parameters of our method were fixed throughout the experiments. For the other compared methods, we used the parameter values defined by authors in the original papers.

\subsection{Ablation study}
In this section, we perform an ablation analysis on OTB-100 to examine the contribution of individual components of our framework. We evaluate three variations of our tracker by comparison to the complete version of our algorithm. The four versions are denoted as follows.
\begin{itemize}
\item \texttt{complete-version}: The complete version of our tracker as described in algorithm \ref{alg:1}.   
\item \texttt{no-fine-prediction}: We limit our appearance modeling to deep features. Target localisation is also limited to the coarse prediction step. In other words, we eliminate the fine prediction step and select the candidate region with the highest likelihood $P_i$ as the final tracking status (see Eq. \ref{eq:2}). 
\item \texttt{no-update}: We eliminate the update mechanism and limit the target appearance modeling to features from the first frame.
\item \texttt{aggressive-update}: Both coarse and fine classifiers are automatically updated at each iteration, without evaluating tracking status (i.e. we do not use quality indicators and minimum quality thresholds). 
\end{itemize}

\begin{figure}[h!]
\centering
\includegraphics[width=0.49\textwidth]{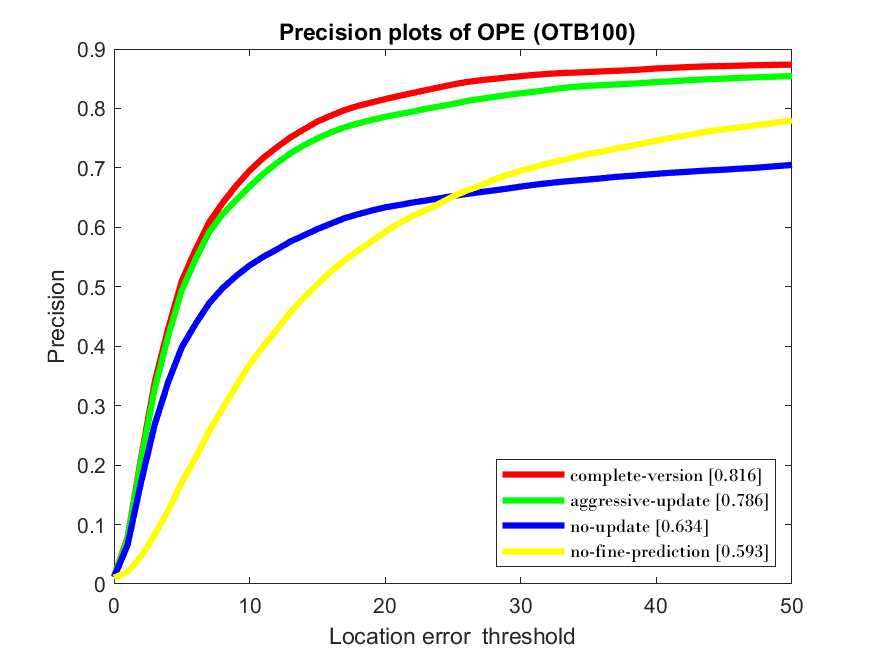}
\includegraphics[width=0.49\textwidth]{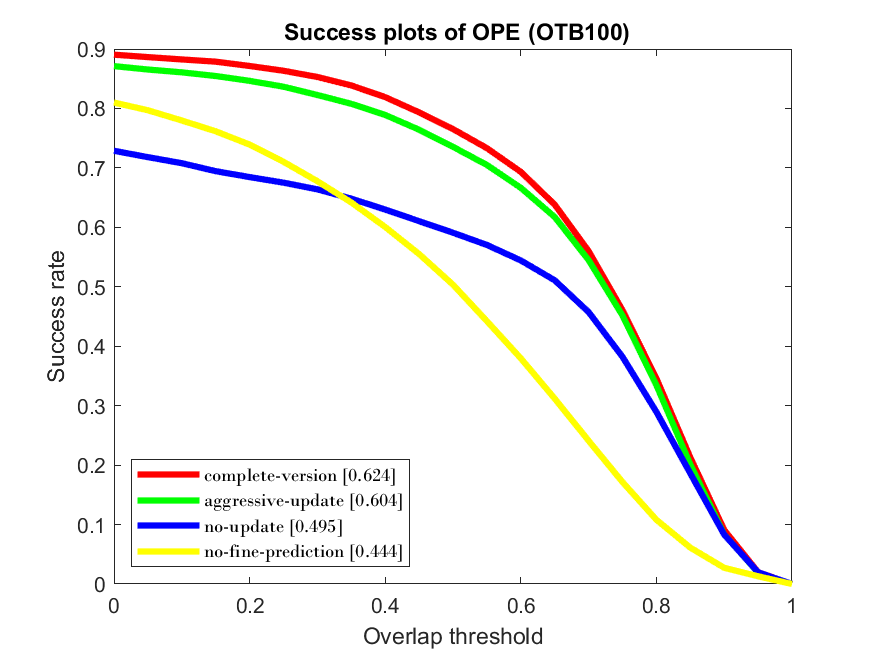}
\caption{ Performance evaluation of different versions of our tracker.}
\label{fig:ablation}
\end{figure}

Figure \ref{fig:ablation} shows precision and success plots on OTB100. It can be clearly observed that the complete implementation of our tracker outperforms all the other versions. The removal of the fine search component from our pipeline (\texttt{no-fine-prediction}) results in a drastic decrease for both precision and success measures. In fact, the convolutional features learned from deep CNN layers are not sufficient for precise localisation, as they do not capture spatial details of the object. Using DCF for subsequent fine prediction allows to considerably improve precision and success respectively by 22\% and 18\%.

Regarding the update procedure, figure \ref{fig:ablation} shows that the complete removal of the update mechanism (\texttt{no-update}) causes a decrease of about 18\% in precision and 13\% in success measures. Furthermore, we can see that the aggressive update strategy (\texttt{aggressive-update}) does not achieve optimal performance either. These results confirm our initial assumption, stating that the proposed update control mechanism is efficient for handling appearance change, while avoiding model contamination. 

\subsection{Comparison with State-of-the Art Trackers}
\begin{figure}[t!]
\centering
\includegraphics[width=0.4\textwidth]{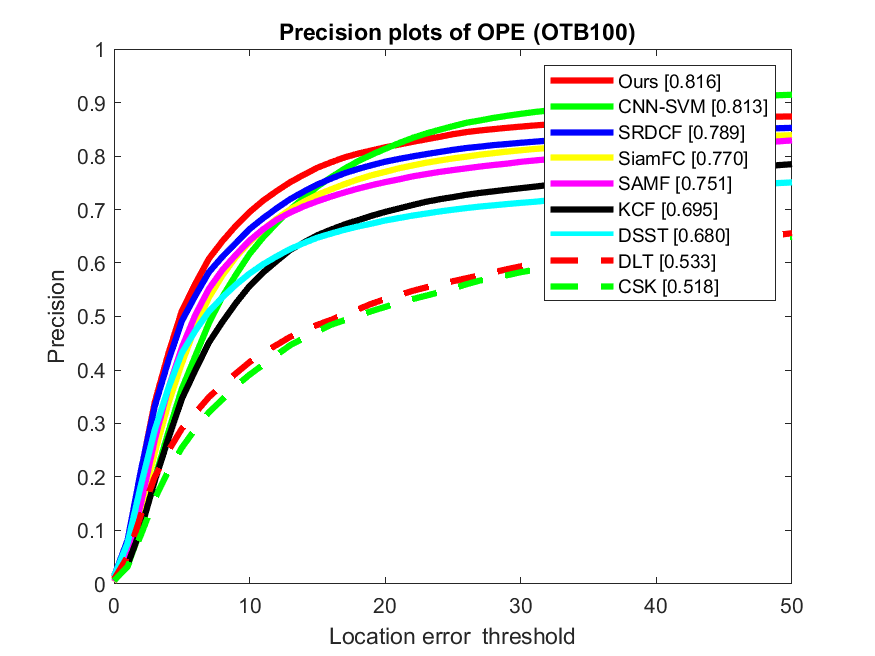}
\includegraphics[width=0.4\textwidth]{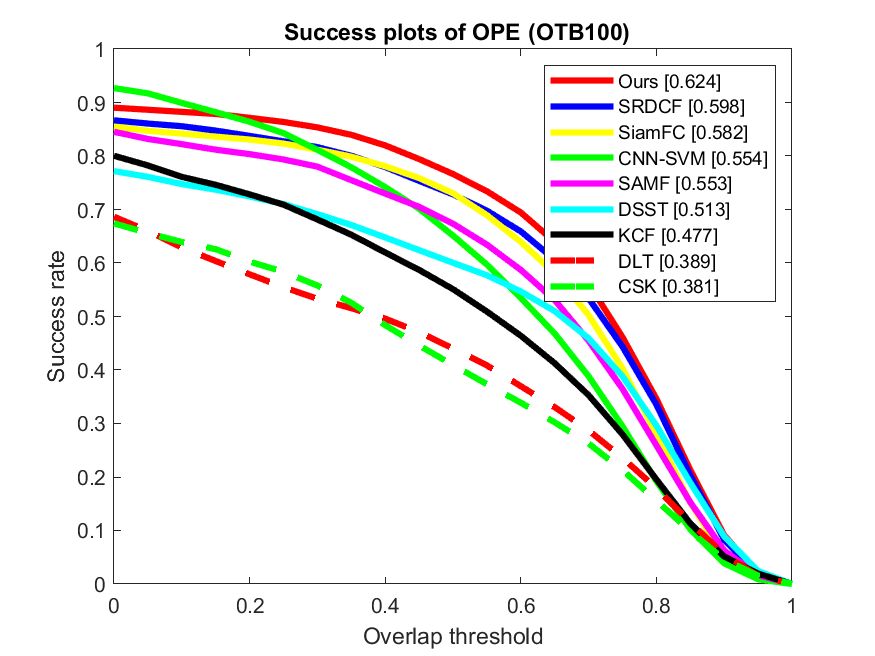}
\includegraphics[width=0.4\textwidth]{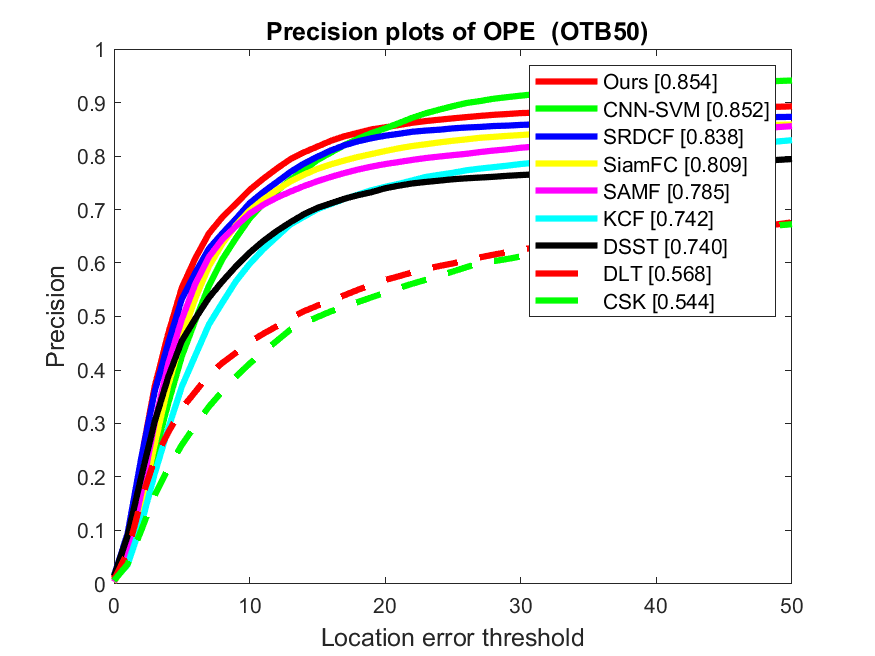}
\includegraphics[width=0.4\textwidth]{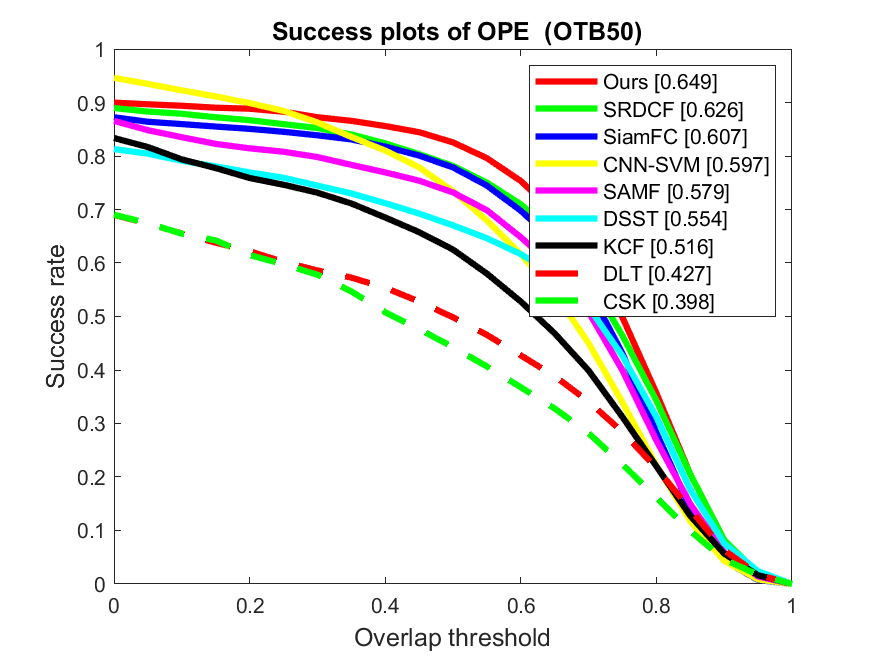}
\caption{Precision and success plots on OTB100 and  OTB50 benchmarks using one-pass evaluation (OPE). The legend of precision plots shows the ranking of the compared trackers based on precision scores at a distance threshold of 20 pixels. The legend of success plots shows a ranking based on the area under-the-curve score.}
\label{fig:eval}
\end{figure}

\textbf{Quantitative evaluation: } We compared our algorithm to several state-of-the-art trackers on OTB100 and OTB50 datasets, which respectively consists of 100 and 50 fully annotated videos with various challenging attributes. Figure \ref{fig:eval} shows the results under one-pass evaluation (OPE), using the distance precision rate and overlap success rate. The overall evaluation shows that our tacker achieved the best performance on both datasets. In particular, we outperformed the CNN-based tracker CNN-SVM \cite{cnnsvm} and the DCF-based tracker SRDCF \cite{srdcf}. Furthermore, the superiority of our algorithm with respect to DCF-based and CNN-based trackers demonstrates that the proposed coarse-to-fine combination of the two approaches allows to improve tracking.

\begin{figure}[htbp]
\includegraphics[scale=0.2]{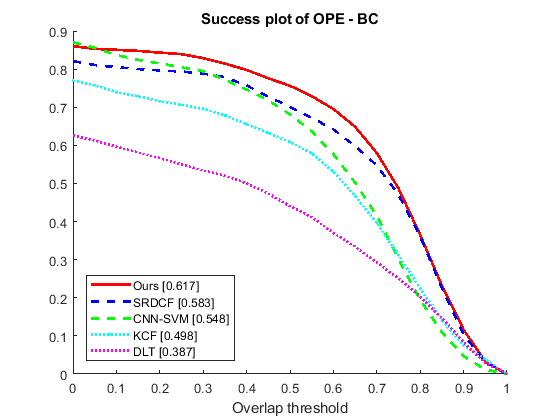}
\includegraphics[scale=0.2]{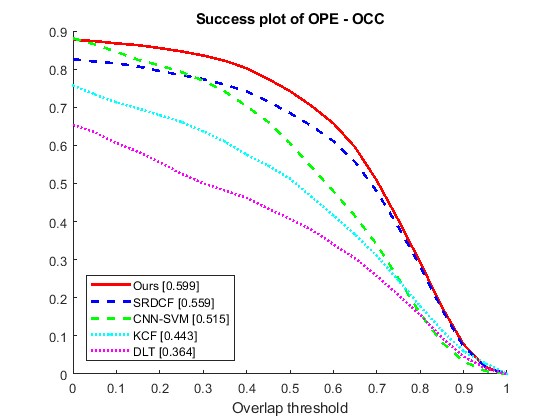}
\includegraphics[scale=0.2]{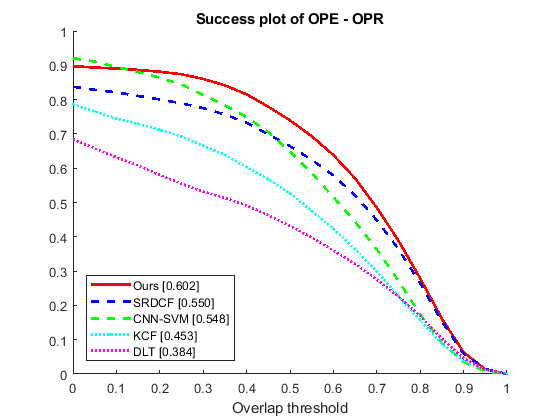}
\includegraphics[scale=0.2]{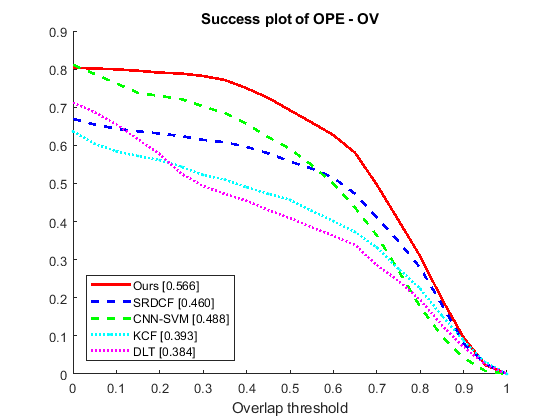}
\includegraphics[scale=0.2]{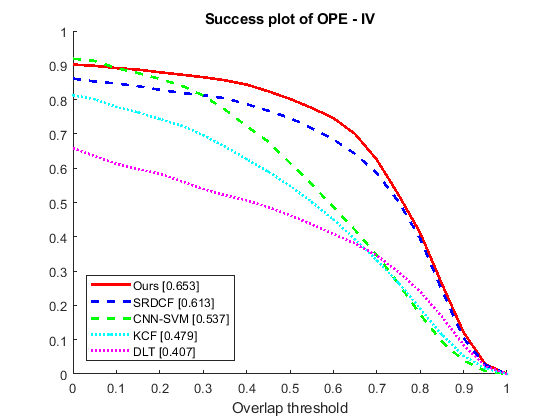}
\includegraphics[scale=0.2]{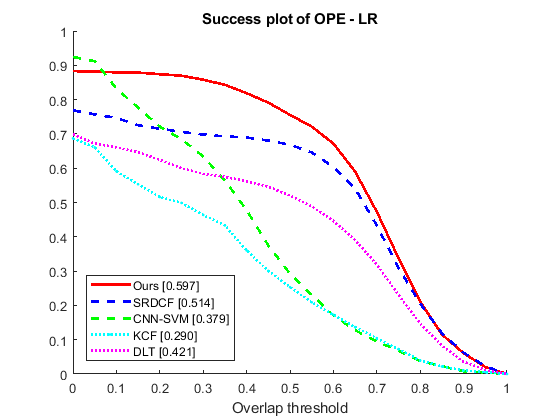}
\includegraphics[scale=0.2]{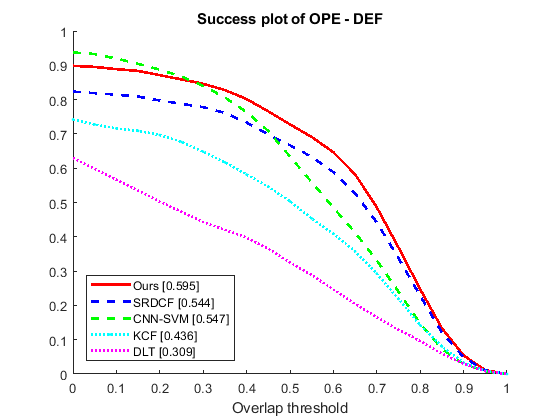}
\includegraphics[scale=0.2]{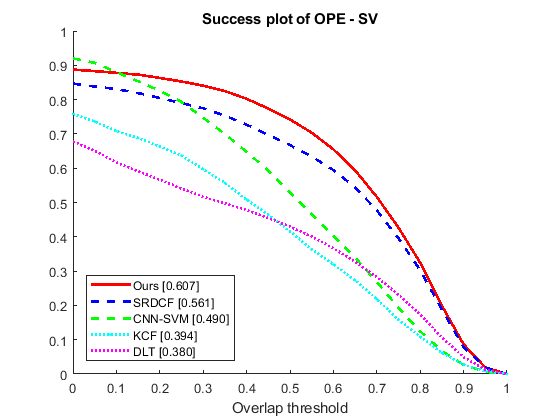}
\caption{ The Success plots on OTB100 for eight attributes representing the challenging aspects in VOT:  background clutter (BC), occlusion (OCC), out-of-plane rotation (OPR), out-of-view (OV), illumination variations (IV), low resolution (LR), deformation (DEF), scale variation (SV).}
\label{fig:attr}
\end{figure}

\textbf{Attribute-based evaluation:} We evaluated the performance of our tracker in different challenging situations. Figure \ref{fig:attr} shows that our tracker outperforms SRDCF \cite{srdcf}, CNN-SVM \cite{cnnsvm}, KCF \cite{kcf}, and DLT \cite{wang2013} on the majority of challenging situations. This evaluation underlines the ability of the proposed tracker to handle all tracking difficulties that generally require high-level semantic understanding of object appearance. In particular, our coarse appearance modeling is proved to be efficient in handling appearance variation caused by out-of-plane rotations, illumination variations, and deformations.

On the other hand, the advantage of using low-level features for fine prediction is illustrated in the background clutter (BC) curve. It is noteworthy that the BC attribute in the OTB is often manifested by the presence of other objects with similar appearance near the target (distractors). In this situation, the two best scores were achieved respectively by our method and SRDCF, as both of them share an important discriminative aspect (the DFC component), which  performs favorably in presence of distractors. Our tracker also effectively deals with the out-of-view and the occlusion problems, where the target is partially or totally invisible during a period of time. Such situations lead to a decrease in the tracking quality indicator $I_t$ (see Eq. \ref{eq:5}), which prevents the model of being contaminated with features from the background. 

\section{Conclusion}
We proposed an effective coarse-to-fine approach for integrating deep features and correlation filters within the tracking framework. First, we exploit the generalization ability of CNNs to coarsely predict target translation between subsequent frames. We then capitalize on the detailed feature representation and the discriminative power of DCFs to obtain a precise location of the target. Once the target is located, appearance model adaptation is carried out through an update control mechanism, which allows the tracker to learn appearance change while avoiding  model drift. The performed experiments demonstrate the efficiency of our approach, indicating improved performances compared to both CNN and DCF-based trackers. Such results confirm that our approach for combining CNNs and DCFs represents a promising direction for developing more advanced and robust trackers.  


%
%
%

 \bibliographystyle{splncs04}
 \bibliography{mybibliography}

\begin{thebibliography}{10}
\providecommand{\url}[1]{\texttt{#1}}
\providecommand{\urlprefix}{URL }
\providecommand{\doi}[1]{https://doi.org/#1}

\bibitem{siamfc}
Bertinetto, L., Valmadre, J., Henriques, J.F., Vedaldi, A., Torr, P.H.:
  Fully-convolutional siamese networks for object tracking. In: ECCV. pp.
  850--865. Springer (2016)

\bibitem{mosse}
Bolme, D.S., Beveridge, J.R., Draper, B.A., Lui, Y.M.: Visual object tracking
  using adaptive correlation filters. CVPR pp. 2544--2550 (2010)

\bibitem{Chi2017DualDN}
Chi, Z., Li, H., Lu, H., Yang, M.H.: Dual deep network for visual tracking. TIP
   \textbf{26},  2005--2015 (2017)

\bibitem{cui2016}
Cui, Z., Xiao, S., Feng, J., Yan, S.: Recurrently target-attending tracking.
  CVPR pp. 1449--1458 (2016)

\bibitem{eco}
Danelljan, M., Bhat, G., Khan, F.S., Felsberg, M.: Eco: Efficient convolution
  operators for tracking. CVPR pp. 6931--6939 (2017)

\bibitem{Danelljan_2019_CVPR}
Danelljan, M., Bhat, G., Khan, F.S., Felsberg, M.: Atom: Accurate tracking by
  overlap maximization. In: CVPR (June 2019)

\bibitem{Danelljan2014AccurateSE}
Danelljan, M., H{\"a}ger, G., Khan, F.S., Felsberg, M.: Accurate scale
  estimation for robust visual tracking. In: BMVC (2014)

\bibitem{deepsrdcf}
Danelljan, M., H{\"a}ger, G., Khan, F.S., Felsberg, M.: Convolutional features
  for correlation filter based visual tracking. ICCV Workshops pp. 621--629
  (2015)

\bibitem{srdcf}
Danelljan, M., H{\"a}ger, G., Khan, F.S., Felsberg, M.: Learning spatially
  regularized correlation filters for visual tracking. ICCV pp. 4310--4318
  (2015)

\bibitem{ccot}
Danelljan, M., Robinson, A., Shahbaz~Khan, F., Felsberg, M.: Beyond correlation
  filters: Learning continuous convolution operators for visual tracking. In:
  ECCV (2016)

\bibitem{ImageNet}
Deng, J., Dong, W., Socher, R., Li, L.J., Li, K., Fei-Fei, L.: Imagenet: A
  large-scale hierarchical image database. CVPR pp. 248--255 (2009)

\bibitem{GCT}
Gao, J., Zhang, T., Xu, C.: Graph convolutional tracking. CVPR pp. 4644--4654
  (2019)

\bibitem{he2018twofold}
He, A., Luo, C., Tian, X., Zeng, W.: A twofold siamese network for real-time
  object tracking. CVPR pp. 4834--4843 (2018)

\bibitem{kcf}
Henriques, J.F., Caseiro, R., Martins, P., Batista, J.: High-speed tracking
  with kernelized correlation filters. TPAMI  \textbf{37},  583--596 (2015)

\bibitem{Henriques2012ExploitingTC}
Henriques, J.F., Caseiro, R., Martins, P., Batista, J.P.: Exploiting the
  circulant structure of tracking-by-detection with kernels. In: ECCV (2012)

\bibitem{cnnsvm}
Hong, S., You, T., Kwak, S., Han, B.: Online tracking by learning
  discriminative saliency map with convolutional neural network. In: ICML
  (2015)

\bibitem{kefi2019novel}
Kefi-Fatteh, T., Ksantini, R., Ka{\^a}niche, M.B., Bouhoula, A.: A novel
  incremental one-class support vector machine based on low variance direction.
  Pattern Recognition  \textbf{91},  308--321 (2019)

\bibitem{siamrpn}
Li, B., Wu, W., Wang, Q., Zhang, F., Xing, J., Yan, J.: Siamrpn++: Evolution of
  siamese visual tracking with very deep networks. CVPR pp. 4277--4286 (2019)

\bibitem{samf}
Li, Y., Zhu, J.: A scale adaptive kernel correlation filter tracker with
  feature integration. In: ECCV Workshops (2014)

\bibitem{Li2018MultiBranchSN}
Li, Z., Bilodeau, G.A., Bouachir, W.: Multi-branch siamese networks with online
  selection for object tracking. In: ISVC (2018)

\bibitem{Ma2015HierarchicalCF}
Ma, C., Huang, J.B., Yang, X., Yang, M.H.: Hierarchical convolutional features
  for visual tracking. ICCV pp. 3074--3082 (2015)

\bibitem{Ma2016WhenCF}
Ma, C., Xu, Y., Ni, B., Yang, X.: When correlation filters meet convolutional
  neural networks for visual tracking. SPL  \textbf{23},  1454--1458 (2016)

\bibitem{Ma2015LongtermCT}
Ma, C., Yang, X., Zhang, C., Yang, M.H.: Long-term correlation tracking. CVPR
  pp. 5388--5396 (2015)

\bibitem{Qi2016HedgedDT}
Qi, Y., Zhang, S., Qin, L., Yao, H., Huang, Q., Lim, J., Yang, M.H.: Hedged
  deep tracking. CVPR pp. 4303--4311 (2016)

\bibitem{vggnet}
Simonyan, K., Zisserman, A.: Very deep convolutional networks for large-scale
  image recognition. CoRR  \textbf{abs/1409.1556} (2015)

\bibitem{touil2019hierarchical}
Touil, D.E., Terki, N., Medouakh, S.: Hierarchical convolutional features for
  visual tracking via two combined color spaces with svm classifier. SIVP
  \textbf{13}(2),  359--368 (2019)

\bibitem{Wang2020TrackingBI}
Wang, G., Luo, C., Sun, X., Xiong, Z., Zeng, W.: Tracking by instance
  detection: A meta-learning approach. In: CVPR. pp. 6288--6297 (2020)

\bibitem{Wang_2019_CVPR}
Wang, G., Luo, C., Xiong, Z., Zeng, W.: Spm-tracker: Series-parallel matching
  for real-time visual object tracking. CVPR pp. 3638--3647 (2019)

\bibitem{wang2013}
Wang, N., Yeung, D.Y.: Learning a deep compact image representation for visual
  tracking. In: NIPS (2013)

\bibitem{Wu2013OnlineOT}
Wu, Y., Lim, J., Yang, M.H.: Online object tracking: A benchmark. CVPR pp.
  2411--2418 (2013)

\bibitem{Wu2015ObjectTB}
Wu, Y., Lim, J., Yang, M.H.: Object tracking benchmark. TPAMI  \textbf{37},
  1834--1848 (2015)

\end{thebibliography}

\end{document}